\documentclass{article}

\usepackage{arxiv}

\usepackage[utf8]{inputenc} 
\usepackage[T1]{fontenc}    

\usepackage{nicefrac}       
\usepackage[numbers, sort&compress]{natbib}
\usepackage{doi}
\usepackage{amsmath}

\usepackage{graphicx}
\usepackage{hyperref}       
\usepackage{url}            
\usepackage{booktabs}       
\usepackage{nicefrac}       
\usepackage{microtype}      
\usepackage{lipsum}
\usepackage{multirow}
\usepackage{comment}

\usepackage{amsfonts,amssymb,bm}
\usepackage{mathtools}
\usepackage{textcomp}
\usepackage{multicol}
\usepackage{bbm}
\usepackage{hyperref}
\usepackage{placeins}
\usepackage{chngpage}
\usepackage{caption}
\usepackage{subcaption}
\usepackage{multicol}
\usepackage{multirow}
\usepackage{booktabs}
\usepackage{adjustbox}
\usepackage{algorithmicx}
\usepackage{algorithm}
\usepackage{algpseudocode}
\algrenewcommand\algorithmicindent{1.0em}

\usepackage{float}

\title{Hybrid Motion Planning with Deep Reinforcement Learning for Mobile Robot Navigation}

\date{}

\newif\ifuniqueAffiliation
\uniqueAffiliationtrue

\ifuniqueAffiliation 
\author{
    {Yury Kolomeytsev} \\
	Department of Computational Mathematics and Cybernetics\\
	Lomonosov Moscow State University\\
	\texttt{yury.kolomeytsev@gmail.com} \\
	\And
	{Dmitry Golembiovsky} \\
	Department of Computational Mathematics and Cybernetics\\
	Lomonosov Moscow State University\\
}
\else
\usepackage{authblk}

\author[1]{%
	Yury Kolomeytsev\thanks{\texttt{yury.kolomeytsev@gmail.com}}}%
\author[1,2]{%
	Dmitry Golembiovsky\thanks{\texttt{}}}%
\affil[1]{Department of Computational Mathematics and Cybernetics}
\affil[2]{Department of Computational Mathematics and Cybernetics}
\fi

\renewcommand{\shorttitle}{Hybrid Motion Planning with Deep Reinforcement Learning}

\hypersetup{
pdftitle={Hybrid Motion Planning with Deep Reinforcement Learning for Mobile Robot Navigation},
pdfsubject={},
pdfauthor={Yury Kolomeytsev, Dmitry Golembiovsky},
pdfkeywords={Robotics, Reinforcement learning, Collision avoidance, Deep learning, Robot navigation, Motion planning},
}

\begin{document}
\maketitle

\begin{abstract} Autonomous mobile robots operating in complex, dynamic environments face the dual challenge of navigating large-scale, structurally diverse spaces with static obstacles while safely interacting with various moving agents. Traditional graph-based planners excel at long-range pathfinding but lack reactivity, while Deep Reinforcement Learning (DRL) methods demonstrate strong collision avoidance but often fail to reach distant goals due to a lack of global context. We propose \emph{Hybrid Motion Planning with Deep Reinforcement Learning} (HMP-DRL), a hybrid framework that bridges this gap. Our approach utilizes a graph-based global planner to generate a path, which is integrated into a local DRL policy via a sequence of checkpoints encoded in both the state space and reward function. To ensure social compliance, the local planner employs an entity-aware reward structure that dynamically adjusts safety margins and penalties based on the semantic type of surrounding agents. We validate the proposed method through extensive testing in a realistic simulation environment derived from real-world map data. Comprehensive experiments demonstrate that HMP-DRL consistently outperforms other methods, including state-of-the-art approaches, in terms of key metrics of robot navigation: success rate, collision rate, and time to reach the goal. Overall, these findings confirm that integrating long-term path guidance with semantically-aware local control significantly enhances both the safety and reliability of autonomous navigation in complex human-centric settings.

\end{abstract}


\section{Introduction}

Mobile robots are rapidly transitioning from controlled industrial settings to complex, human-centric environments. Robots designed to assist humans are increasingly being deployed across a wide range of sectors, including manufacturing and industrial automation, agriculture, space exploration, healthcare, domestic assistance, logistics and delivery, and facility and public-space services \cite{Siegwart2011,FRAGAPANE2021405,AutonomousRobots}.
Autonomous mobile robots combine a mobile platform, sensors, and onboard intelligence to adapt in real time to dynamic environments, rather than depending on fixed infrastructure like automated guided vehicles (AGVs).
A central capability underlying these systems is autonomous navigation, which allows a robot to operate independently, perceive the environment, and act safely to achieve specified goals without operator supervision or control.

Typical applications of autonomous robots include material handling in warehouses and factories; hospital logistics (e.g., transporting laboratory samples, blood components, and medications); service robots in commercial buildings and hotels for indoor delivery and visitor guidance; autonomous cleaning; security patrol and inspection; and last-mile urban delivery of groceries, meals, and parcels \cite{AutonomousRobots,drew2021multi,KRUSE20131726}.

The core challenge in such deployments is twofold: traversing large-scale, structurally complex environments (global planning) while safely interacting with diverse dynamic agents (local planning and collisions avoidance).
Consider a sidewalk delivery robot \cite{SADR} operating in an urban setting: it must travel up to several kilometers, traversing diverse surfaces such as sidewalks and crosswalks, while accounting for static obstacles like buildings, bus stops, flowerbeds and curbs. Traditional search-based planners (e.g., A*, Dijkstra) excel at finding optimal paths over these long distances using static maps but struggle to react fluidly to unmapped, moving entities.

Conversely, Deep Reinforcement Learning (DRL) methods have demonstrated superior performance in local collision avoidance and crowd navigation \cite{CADRL,SARL,EBCADRL}. However, pure DRL approaches typically lack global context.
When applied to large environments with complex geometry, they often fail to reach distant goals, getting trapped in local minima due to the absence of global guidance.

Furthermore, safe navigation requires more than just treating obstacles as generic geometric shapes. A robot must navigate in a socially compliant manner, accounting for the semantic diversity of the environmental entities around it. A child, a bicyclist, an adult, and a concrete wall all impose different safety constraints and motion patterns. Treating a vulnerable pedestrian with the same safety margin as a static obstacle is insufficient for socially compliant navigation.

To address these challenges, we introduce \emph{Hybrid Motion Planning with Deep Reinforcement Learning} (HMP-DRL), a novel hybrid method that bridges the gap between global path planning and local motion planning with entity-aware collision avoidance. Our approach leverages a global path planner to provide long-term guidance through complex static environments (handling curbs, walls, and restricted zones), while a local DRL policy handles immediate interaction with dynamic agents and static entities. This decomposition allows the robot to navigate kilometers without losing its way, while simultaneously exhibiting sophisticated, entity-aware collision avoidance to ensure safe navigation.

Our main contributions are as follows:
\begin{itemize}
    \item We propose a hybrid motion planning method that combines graph-based global search with deep RL-based local planning. This hybrid structure enables the robot to utilize map information for reaching distant goals while maintaining the reactivity required for dynamic obstacle avoidance.
    \item We design a reward function that encourages the robot to follow the global path and minimize travel time. The function uses semantic awareness by penalizing the robot differently depending on the type of agent or obstacle it encounters. Specifically, the required safe distance and collision penalties are dynamically adjusted based on the entity type, resulting in safer and more socially compliant behavior.
    \item We introduce a training and testing environment designed to mimic realistic outdoor settings with various static and dynamic obstacles. Our experimental results demonstrate that this new approach consistently outperforms standard navigation techniques, including state-of-the-art methods.
\end{itemize}

\section{Related Work}\label{sec:relwork}

Robot navigation research has evolved from traditional rule-based methods and classical planning algorithms to learning-based approaches, with recent work focusing on crowd navigation and hybrid architectures that combine global and local planning.

Classical approaches such as the Social Force Model (SFM) \cite{social_force} and Optimal Reciprocal Collision Avoidance (ORCA) \cite{ORCA} use hand-crafted interaction rules for collision avoidance. SFM applies attractive and repulsive forces but is prone to local minima; ORCA uses velocity constraints under reciprocal assumptions, expecting all agents to cooperate equally in avoiding collisions—an assumption rarely met in real-world settings with unpredictable pedestrians. Both methods can produce short-sighted or unnatural behavior and struggle to generalize across diverse social environments.

Learning-based methods have gained significant attention as alternatives to hand-crafted rules. Imitation learning approaches derive navigation strategies from demonstrations of desired behaviors. For example, \cite{navigation_raw_depth_inputs_2018} combines behavior cloning with generative adversarial imitation learning (GAIL) to develop a policy that operates directly on raw depth images, bypassing explicit pedestrian tracking. \cite{Kretzschmar_2016} applies maximum entropy inverse RL to model human navigation as a mixture distribution capturing both discrete decisions (e.g., passing left or right) and continuous trajectory variance. However, these methods depend heavily on the quality and scale of demonstrations, which can be resource-intensive to collect and may limit generalization to novel scenarios.

Deep reinforcement learning (DRL) offers an alternative approach by treating navigation as a sequential decision-making problem. CADRL \cite{CADRL} formulates decentralized multi-agent collision avoidance by training a value network that estimates time-to-goal from the joint state of the robot and its neighbors, achieving faster navigation than ORCA while remaining real-time implementable. SA-CADRL \cite{socially_aware_2017} extends this idea with norm-inducing rewards that penalize violations of social conventions (e.g., passing on the wrong side) and a symmetric network architecture with weight-sharing and max-pooling to handle variable numbers of agents, trading some efficiency for socially compliant behavior. To better scale to dense crowds, LSTM-RL \cite{LSTM-RL} uses recurrent networks to encode an arbitrary number of neighboring agents, significantly reducing failure rates in crowded scenarios. SARL \cite{SARL} further improves crowd modeling through an attention-based value network that jointly reasons about human-robot and human-human interactions, enhancing anticipation of crowd dynamics. More recently, \cite{Xue2024} proposes a spatial-temporal value network and a hazard-aware reward structure that explicitly reasons about pedestrian speeds and hazardous regions to improve safety and comfort. Recently, EB-CADRL \cite{EBCADRL} introduced entity-type awareness into the state representation and reward function, enabling differentiated safety margins for various agent types. The method demonstrated superior safety for vulnerable agents in dense crowds and achieved significant training speedups through parallelized experience collection.

A significant limitation of pure DRL methods is that they focus primarily on local collision avoidance and are typically trained and evaluated in small, bounded environments. When applied to large-scale maps or long-range navigation tasks, these methods fail to reach distant goals, often getting stuck or exhibiting suboptimal behavior due to the absence of global path guidance. Furthermore, they neglect static environmental structures such as buildings, walls, and terrain boundaries. To bridge this gap, recent research has explored hybrid approaches that combine classical path planning with learning-based local control.

DRL-based hybrid methods integrate global planners with learned local policies. SARL* \cite{SARL_STAR} extends SARL by using Dijkstra's algorithm for dynamic local goal setting and a map-based safe action space; however, it rebuilds paths every time step (computationally inefficient) and relies on rule-based action filtering rather than learned obstacle avoidance. G2RL \cite{G2RL} incorporates a static A* path as a guidance channel into the DRL state space, but the fixed global path becomes problematic when obstacles block the path, and the coarse discretization (4-directional actions on $200 \times 200$ grids) limits real-world applicability. Similarly, \cite{Yurtsever2020} integrates A* with a DQN agent for automated driving, demonstrating faster convergence, but lacks pedestrian awareness and uses discrete actions resulting in jerky control. GDAE \cite{GDAE} combines frontier-based waypoint selection with TD3 for unknown environments but does not utilize a global path and lacks dynamic obstacle handling. VDPF-TD3 \cite{VDPF_TD3} uses TD3 to optimize the balance between modified potential field forces, effectively escaping local optima, but relies on geometric obstacle representations and lacks real-world validation.

Classical hybrid planners combine global path planning with reactive local control. Zhong et al. \cite{SafeAStar_DWA} propose Safe A* with risk-aware costs combined with an Adaptive Window Approach, while IAB-A* \cite{IAB_AStar} introduces bidirectional search with B-spline smoothing integrated with DWA. Both methods achieve computational efficiency but depend heavily on manually tuned parameters and do not differentiate between obstacle types. The RC-Map approach \cite{RCMap} uses rectangular decomposition for fast topological planning (10ms vs 740ms for grid methods) but produces sub-optimal paths and struggles with non-orthogonal environments.

Sampling-based hybrids improve path search efficiency. RRTstar-RL \cite{RRTstar_RL} replaces uniform RRT* sampling with a learned probability distribution, achieving 28\% shorter paths and 2x speedup, but requires extensive training data and lacks mechanisms for dynamic obstacle avoidance.



Our HMP-DRL approach addresses these limitations by (i) computing a single global path per episode and using its checkpoints in both the reward and the network input to provide path guidance at low cost, (ii) learning collision avoidance directly through the reward with a fine-grained discrete action space that still enables smooth motion, and (iii) encoding entity types in the state and reward to achieve socially compliant navigation with type-dependent safety margins.

\section{Problem Formulation}\label{sec:problem_formulation}

We address the problem of autonomous robot navigation towards a goal in dynamic environments populated with various entities, including dynamic agents and static obstacles. This problem can be formulated as a Markov decision process (MDP).
The robot must navigate in a socially compliant manner, avoiding collisions while reaching its goal. We define the state of each agent (robot or environmental entity) in terms of observable and unobservable components:

\begin{itemize}
\item Observable states: position $\mathbf{p} = [p_x, p_y]$, velocity $\mathbf{v} = [v_x, v_y]$, and radius $r$.
\item Unobservable states: goal position $\mathbf{g} = [g_x, g_y]$, preferred speed $v_{\text{pref}}$, and heading angle $\theta$.
\end{itemize}

While acting in the environment, the robot knows its observable and unobservable states, as well as the observable states of all entities. We use the robot-centric parameterization described in \cite{CADRL} and \cite{LSTM-RL}, where the robot is positioned at the origin and the x-axis is aligned with the robot's goal direction. Denote $e_i$ as a type of entity $i$. The robot’s state and the $i$-th entity's observable state at time $t$ could be represented as:
\begin{equation}
\begin{aligned}
    \mathbf{s}_t^r &= [d_g, v_x, v_y, r, v_{\text{pref}}, \theta], \\
    \mathbf{s}_t^{i o} &= [p_x^i, p_y^i, v_x^i, v_y^i, r^{i}, d^i, r^{i} + r, e_i],
\end{aligned}
\label{eq:states}
\end{equation}
where \(d_g = \|\mathbf{p}_t - \mathbf{g}\|_2\) is the distance from the robot to the goal, and \(d^i = \|\mathbf{p}_t - \mathbf{p}_t^i\|_2\) is the distance from the robot to the $i$-th entity.

The robot can acquire its own state and the observable states of other agents at each time step. The joint state at time $t$ is defined by:
\begin{equation}
\mathbf{s}_t^j = [\mathbf{s}_t^r, \mathbf{s}_t^{1 o}, \mathbf{s}_t^{2 o}, \ldots, \mathbf{s}_t^{n o}].
\label{eq:states_2}
\end{equation}

The robot’s velocity \(\mathbf{v}_t\) is determined by the action command \(\mathbf{a}_t\) from the navigation policy, i.e., \(\mathbf{v}_t = \mathbf{a}_t = \pi(\mathbf{s}_t^j)\).

The reward function \(R(\mathbf{s}_t^j, \mathbf{a}_t)\) is defined as a mapping from the current state \(\mathbf{s}_t^j\) and action \(\mathbf{a}_t\) to a scalar value, which represents the immediate benefit of taking action \(\mathbf{a}_t\) in state \(\mathbf{s}_t^j\). Formally:
\begin{equation}
R(\mathbf{s}_t^j, \mathbf{a}_t) = r_t,
\label{eq:reward}
\end{equation}
where \(r_t\) is the reward received at time step \(t\).

Our objective is to find the optimal policy \(\pi^*(\mathbf{s}_t^j)\) that maximizes the expected reward.

The optimal value function is given by \cite{Sutton}:
\begin{equation}
V^*(\mathbf{s}_t^j) = \sum_{\tilde{t}=t}^T \gamma^{\tilde{t} \cdot v_{\text{pref}}} R_{\tilde{t}}(\mathbf{s}_{\tilde{t}}^j, \pi^*(\mathbf{s}_{\tilde{t}}^j)),
\label{eq:optimal_value_func}
\end{equation}
where \(\gamma \in (0, 1)\) is a discount factor and $T$ is the time step at which the episode ends.

Using value iteration, the optimal policy is derived as:

\begin{equation}
\begin{aligned}
& \pi^*\left(\mathbf{s}_t^j\right)=\arg \max _{\mathbf{a}_t \in \mathbf{A}} \biggl[ R\left(\mathbf{s}_t^j, \mathbf{a}_t\right)+\gamma^{\Delta t \cdot v_{\text {pref }}} \cdot \\
& \quad \int_{\mathbf{s}_{t+\Delta t}^j} P\left(\mathbf{s}_{t+\Delta t}^j \mid \mathbf{s}_t^j, \mathbf{a}_t\right) V^*\left(\mathbf{s}_{t+\Delta t}^j\right) d \mathbf{s}_{t+\Delta t}^j \biggr],
\end{aligned}
\label{eq:optimal_policy}
\end{equation}

where $\Delta t$ is time step duration, \(R(\mathbf{s}_t^j, \mathbf{a}_t)\) is the reward function, \(\mathbf{A}\) is the action space, and \(P(\mathbf{s}_{t+\Delta t}^j \mid \mathbf{s}_t^j, \mathbf{a}_t)\) is the transition probability.

\section{Proposed Approach}\label{sec:new_approach}


In this section, we introduce our new approach which we call Hybrid Motion Planning with Deep Reinforcement Learning \mbox{(HMP-DRL)}. HMP-DRL combines a global graph-based path planner with a local deep reinforcement learning policy, enabling long-range navigation through complex environments with static and dynamic obstacles while maintaining entity-aware collision avoidance for safe, socially compliant interaction with dynamic agents.

\subsection{Global Planning}\label{subsec:global}

Global path planning determines an optimal collision-free path from a starting position to a goal within a known or partially known environment. The core idea is to construct a graph representation of the navigable space and then apply a search algorithm to find the best path through this graph. The global planner considers the map and known static obstacles. Dynamic agents and static obstacles not present on the map are unknown at global planning time but are detected by the robot's sensors during navigation and handled by the local planner in real time.

\paragraph{Graph Construction Methods.}
A fundamental aspect of global planning is the construction of a graph that captures the connectivity of free space. Different methods can be used to build such graphs, each with distinct advantages:

\begin{itemize}
\item \textit{Occupancy Grids.}
The environment is discretized into a regular grid of cells, where each cell is marked as occupied or free based on the presence of obstacles. Nodes are placed at the centers of free cells, and edges connect adjacent nodes, enabling grid-based pathfinding algorithms. This method is straightforward, widely used in 2D environments, and integrates seamlessly with sensor data such as occupancy maps.
\item \textit{Navigation Meshes.}
Free space can be partitioned into a set of connected polygons, most commonly triangles or quadrangles, using cell decomposition methods \cite{Latombe2012}. Unlike uniform grids, mesh-based representations such as Irregular Triangular Meshes (ITM) and Varying-size ITMs (VITM) can better conform to obstacle boundaries. In particular, VITM enables local refinement in narrow passages while maintaining coarser cells in open areas, providing a balance between geometric fidelity and computational complexity \cite{Kallmann2005}. Quadrangle (QUAD) representations, although less flexible in approximating irregular or oblique obstacle boundaries, offer a simple and structured topology with predictable neighbor relationships. When an inflation layer is applied to account for robot size, QUAD meshes are generally less sensitive than fixed-size triangular meshes, but they may fail to preserve connectivity in narrow or cluttered environments unless a fine resolution is used. In contrast, varying-size triangular meshes demonstrate greater robustness under inflation, particularly in environments with narrow passages, online map updates, or multiple robots with differing dimensions, making them more suitable for such scenarios \cite{Meysami2022}.
\item \textit{Voronoi Diagrams.}
Given a vectorized map with polygonal obstacle boundaries, a Generalized Voronoi Diagram (GVD) of the free space can be constructed \cite{Choset1997GVG, Aurenhammer1991}. The GVD edges consist of points equidistant to two or more obstacles, forming a sparse skeleton of the navigable space. The Voronoi vertices (where edges meet) serve as nodes, and the edges themselves serve as paths in the planning graph. Paths computed on Voronoi-based graphs locally maximize clearance from obstacles, improving safety in cluttered or narrow environments, albeit often at the cost of increased path length compared to shortest-path methods.
\item \textit{Visibility Graphs.}
In environments with polygonal obstacles, nodes can represent the start, goal, and obstacle vertices, with edges connecting mutually visible pairs \cite{LozanoPerez1979, VG_path_planning_2023}. This yields the shortest path in terms of Euclidean distance but may produce trajectories that pass too close to obstacle corners.
\end{itemize}

\begin{minipage}{0.9\linewidth}

\begin{algorithm}[H]
\caption{A* Algorithm}\label{algorithm_a_star}
\begin{algorithmic}[1]
\Require occupancy map, start node \(s\), goal node \(g\), neighbor function \(\mathcal{N}(\cdot)\), edge cost \(c(\cdot,\cdot)\), heuristic \(h(\cdot)\)
\Ensure shortest path from \(s\) to \(g\), if it exists
\State openSet \(\gets\) \{\(s\)\}, closedSet \(\gets\) \(\emptyset\)
\State cameFrom \(\gets\) empty map
\State \(g\)Score\((s) \gets 0\), \(f\)Score\((s) \gets h(s)\)
\While{openSet is not empty}
	\State \(n \gets \arg\min_{x\in\text{openSet}} f\text{Score}(x)\)
	\If{\(n = g\)}
	\State \Return \Call{ReconstructPath}{cameFrom, n}
	\EndIf
	\State remove \(n\) from openSet; add \(n\) to closedSet
	\For{each neighbor \(m \in \mathcal{N}(n)\)}
	\If{\(m\) in closedSet or grid cell \(m\) is occupied}
		\State \textbf{continue}
	\EndIf
	\State tentative \(\gets g\)Score\((n) + c(n,m)\)
	\If{\(m\) not in openSet}
		\State add \(m\) to openSet
	\ElsIf{tentative \(\geq g\)Score\((m)\)}
		\State \textbf{continue}
	\EndIf
	\State cameFrom\((m) \gets n\)
	\State \(g\)Score\((m) \gets\) tentative
	\State \(f\)Score\((m) \gets g\)Score\((m) + h(m)\)
	\EndFor
\EndWhile
\State \Return failure

\Function{ReconstructPath}{cameFrom, n}
	\State path \(\gets\) [\(n\)]
	\While{cameFrom contains \(n\)}
	\State \(n \gets\) cameFrom\((n)\)
	\State prepend \(n\) to path
	\EndWhile
	\State \Return path
\EndFunction
\end{algorithmic}
\end{algorithm}

\end{minipage}

All four graph construction methods are applicable in our HMP-DRL framework; the choice depends on the available map representation and the desired path characteristics. Occupancy grids are often preferred when maps originate from SLAM or semantic segmentation pipelines due to their simplicity and direct compatibility with sensor data. Navigation meshes provide an efficient representation for large open areas with complex obstacle shapes, enabling smooth paths through convex regions while naturally handling varying terrain costs. Voronoi-based graphs are advantageous when vector maps with polygon boundaries are available and maximum clearance from obstacles is prioritized over path length. Visibility graphs yield the shortest Euclidean paths but may require additional safety margins since trajectories can pass close to obstacle corners.

\paragraph{Search Algorithm.}
We employ the A* algorithm \cite{Hart1968} as the global planner to compute the shortest collision-free path between the start and the goal on the constructed graph. A* combines the exact path cost from the start with a heuristic estimate to the goal, and, given an admissible and consistent heuristic (e.g., Euclidean distance), it is both complete and optimal, typically expanding significantly fewer nodes than Dijkstra's algorithm \cite{Dijkstra1959}. In contrast, Breadth-First Search (BFS) \cite{Cormen_algorithms} is optimal only on unweighted graphs and explores many irrelevant states; Dijkstra's algorithm \cite{Dijkstra1959} is optimal with non-negative costs but lacks heuristic guidance, making it less efficient; incremental planners like D* \cite{D_STAR} efficiently update paths when costs change but add unnecessary implementation complexity and computational overhead not required in our static global scenario \cite{lavalle_2006}.

The operation of the A* algorithm can be described as follows. An open list is maintained, ordered by priority \(f(n)=g(n)+h(n)\), where \(g\) is the path cost from the start to node \(n\), and \(h\) heuristically estimates the cost from \(n\) to the goal. At each step, the node with the minimum \(f\) is expanded, neighbors are relaxed, and the process repeats until the goal is selected; the optimal path is then reconstructed using backpointers. On the grid, we use occupancy checks to avoid static obstacles and a distance heuristic that preserves optimality. In our implementation, we use an 8-connected grid where each cell's neighbor set \(\mathcal{N}(n)\) includes all 8 adjacent cells (4 orthogonal and 4 diagonal neighbors), enabling diagonal movement paths.
For completeness, we provide the detailed pseudocode of the A* procedure in Algorithm \ref{algorithm_a_star}.

\paragraph{Experimental Setup.}
In our experiments, the robot operates outdoors in an urban street environment (e.g., delivery robots, security patrol robots, and cleaning robots). We use the occupancy grid method: the map covers an area of 200\,m by 200\,m and is discretized into cells of 10\,cm by 10\,cm. The occupancy grid is formed from street semantic layers: areas such as buildings, lawns, vacant lots, sports fields, and other non-traversable or restricted zones are marked as occupied (forbidden for movement), while traversable road sections and pedestrian paths are marked as free space.

\begin{figure}[!h]
\center{\includegraphics[width=0.58\linewidth]{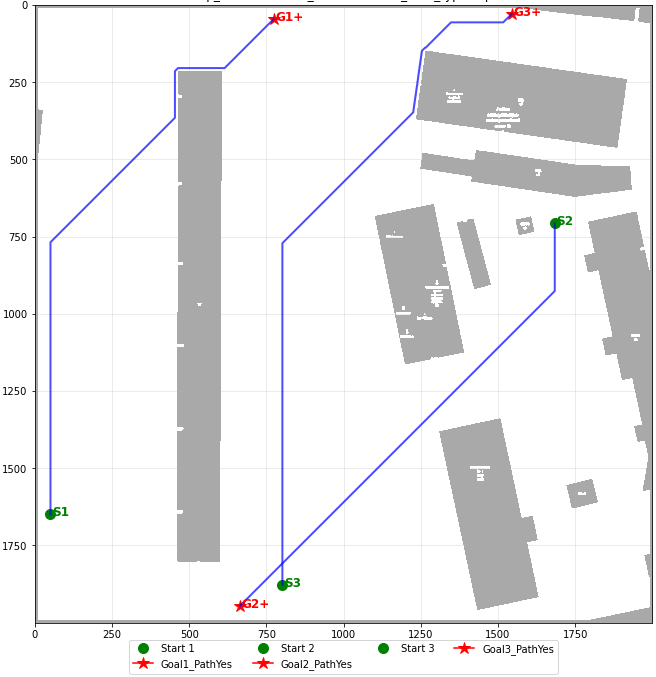}} 
    \caption{Three examples of global paths constructed by the A* algorithm for different start and goal pairs in our experiments. Occupied cells are marked in gray, which in this episode represent residential and non-residential buildings of various sizes and shapes in the city.}
    \label{fig:a_star_paths}
\end{figure}

Examples of paths constructed using the A* algorithm in our experiments are shown in Fig. \ref{fig:a_star_paths}.

\subsection{Local Planning}\label{subsec:local}

Our local planning is based on the EB-CADRL approach described in \cite{EBCADRL}, but with enhancements to incorporate the global path into the model and the reward function.
The local planner considers both static and dynamic obstacles and aims to follow the global path.

To link global planning with local decision-making, we place a set of checkpoints along the global path. The checkpoints are positioned at a distance \(D\) from each other. The first checkpoint is located on the global path at a distance \(D\) from the start. Checkpoints can have different shapes and sizes. In our experiments, we use circular points with a fixed radius. These checkpoints are visible to the local planner and are included in both the reward function and the neural network $V$, encouraging progress along the global path.

\begin{figure}[!h]
\center{\includegraphics[width=0.9\linewidth]{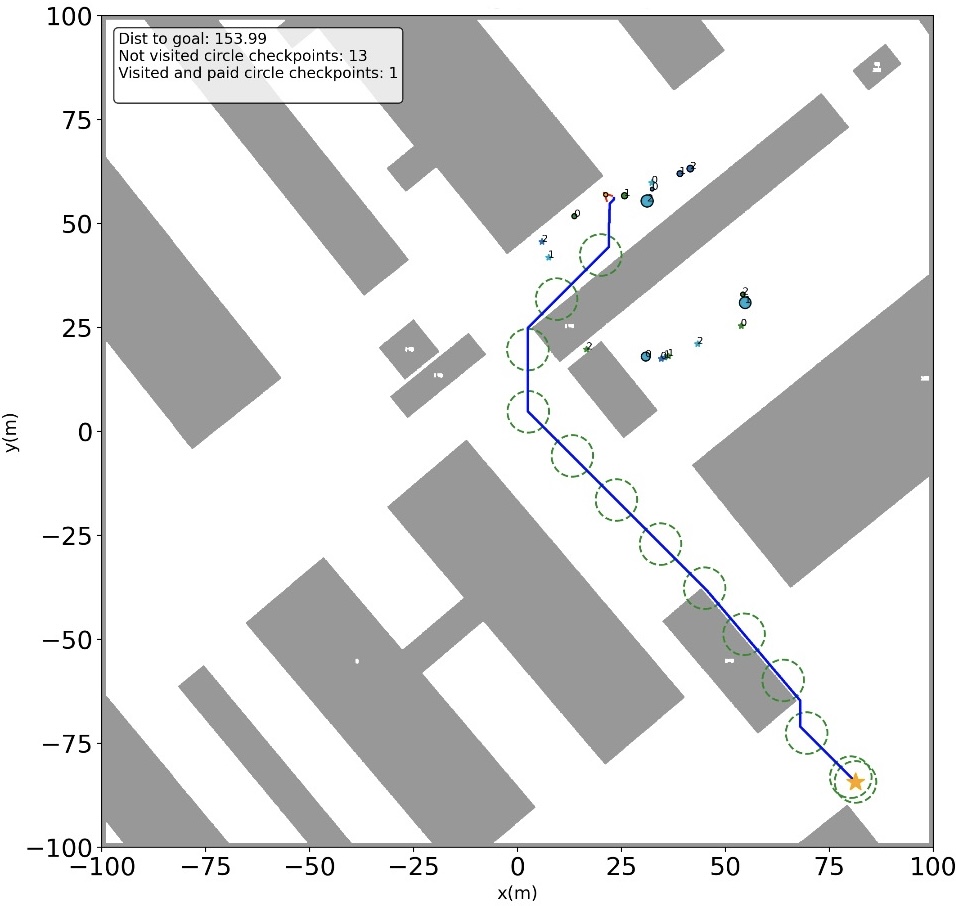}} 
    \caption{Simulator snapshot during the simulation of the HMP-DRL method. The blue polyline is the global path. Green circles are checkpoints. Gray objects are residential and non-residential buildings in the city. The remaining objects are the robot and the dynamic agents surrounding it.}
    \label{fig:checkpoints}
\end{figure}

Fig. \ref{fig:checkpoints} depicts a snapshot of the simulator during an HMP-DRL simulation. In it, we can see how checkpoints (green dashed circles) are positioned along the global path.

\subsection{Reward Function}\label{subsec:reward_function}

Our reward function $R(t)$ comprises three additive components: a terminal reward $R_{\text{terminal}}(t)$ for episode-ending events (goal reached, collision, or timeout), a checkpoint reward $R_{\text{checkpoint}}(t)$ for progress along the global path, and a discomfort penalty $R_{\text{discomfort}}(t)$ for proximity violations.

Before defining these components, we establish notation for measuring robot-entity distances. Consider an environment with $N$ entity types $\mathcal{E} = \{e_1, \ldots, e_N\}$. Let $d(e_i, t)$ denote the minimum distance between the robot and any entity of type $e_i$ over the interval $[t - \Delta t, t]$. We define:
\begin{equation}
i^*(t) = \arg\min_{i \in \{1, \ldots, N\}} d(e_i, t),
\label{eq:closest_entity_index}
\end{equation}
where $i^*(t)$ is the index of the entity type closest to the robot. For brevity, we write $i^*$ instead of $i^*(t)$. The minimum distance to any entity is $d_{\text{min}}(t) = d(e_{i^*}, t)$, and $e_{i^*} \in \mathcal{E}$ is the corresponding entity type. A collision occurs when $d_{\text{min}}(t) \leq 0$.

The robot receives a checkpoint reward $R_{\text{checkpoint}}(t)$ when it is inside a previously unvisited checkpoint along the A* path. Checkpoints may have different shapes and sizes. For example, we could use circular checkpoints, and the robot receives a reward when the Euclidean distance from the robot to the checkpoint's center is less than or equal to the checkpoint radius. Each checkpoint is paid out at most once (no repeated payments).

The function $R_{\text{time}}(t)$ represents the time reward:
\begin{equation}
R_{\text{time}}(t) = 
\begin{cases}
1 & \text{if } t < t_{\text{pref}} \\
\frac{t_{\text{max}} - t}{t_{\text{max}} - t_{\text{pref}}} & \text{if } t_{\text{pref}} \leq t \leq t_{\text{max}} \\
0 & \text{otherwise}
\end{cases}
\label{eq:time_reward}
\end{equation}
where $t_{\text{pref}}$ is the preferred time duration for reaching the goal, $t_{\text{max}}$ is the maximum allowable time to reach the goal.

Let $d_{\text{max}}$ be the maximum distance to the goal, which is the Euclidean distance between the starting position of the robot at time $t = 0$ and the position of the goal. Let $d_g(t)$ be the distance to the goal at time $t$. Then proximity reward can be represented as:
\begin{equation}
\begin{aligned}
R_{\text{prox}}(t) = 1 - \frac{d_g (t)}{d_{\text{max}}}.
\end{aligned}
\label{eq:proximity_reward}
\end{equation}

Define $R_{\text{coll}}(e_i)$ as the penalty associated with collisions involving an entity of type $e_i$, and $R_{\text{success}}$ as the reward for reaching the goal.

Let $d_{\text{disc}}(e_i)$ be the discomfort distance for entity of type $e_i$ and $p_{\text{disc}}(e_i)$ be the discomfort penalty factor for entity of type $e_i$, $i = 1, \ldots, N$.

Our reward function is defined as a sum of three components:
\begin{equation}
R(t) = R_{\text{terminal}}(t) + R_{\text{checkpoint}}(t) + R_{\text{discomfort}}(t).
\label{eq:reward_function}
\end{equation}

The terminal reward handles mutually exclusive end-of-episode events and incorporates the proximity reward for timeout and collision cases:
\begin{equation}
R_{\text{terminal}}(t) =
\begin{cases}
R_{\text{prox}}(t) & \text{if } t \geq t_{\text{max}} \text{ and } \lVert \mathbf{p}_t - \mathbf{g} \rVert_2 > r \\
R_{\text{coll}}(e_{i^*}) + R_{\text{prox}}(t) & \text{else if } d_{\text{min}}(t) \leq 0 \\
R_{\text{success}} + R_{\text{time}}(t) & \text{else if } \lVert \mathbf{p}_t - \mathbf{g} \rVert_2 \leq r \\
0 & \text{otherwise}
\end{cases}
\label{eq:terminal_reward}
\end{equation}
where $r$ is the robot's radius. The proximity reward $R_{\text{prox}}(t)$ is added to timeout and collision cases to provide a learning signal proportional to the robot's progress toward the goal, even when the episode ends unsuccessfully.

The checkpoint reward encourages progress along the global path:
\begin{equation}
R_{\text{checkpoint}}(t) =
\begin{cases}
R_{\text{cp}} & \text{if } \mathbf{p}_t \in \mathcal{C}_k \text{ and } k \text{ is unvisited} \\
0 & \text{otherwise}
\end{cases}
\label{eq:checkpoint_reward}
\end{equation}
where $\mathcal{C}_k$ denotes the region of checkpoint $k$ (e.g., a circle, rectangle, or line segment with a tolerance width), and $R_{\text{cp}}$ is the checkpoint reward value. Each checkpoint yields a reward at most once.

The discomfort reward penalizes proximity to entities:
\begin{equation}
R_{\text{discomfort}}(t) =
\begin{cases}
(d(e_{i^*}, t) - d_{\text{disc}}(e_{i^*})) \cdot p_{\text{disc}}(e_{i^*}) \cdot \Delta t & \text{if } d(e_{i^*}, t) < d_{\text{disc}}(e_{i^*}) \text{ and } d_{\text{min}}(t) > 0 \\
0 & \text{otherwise}
\end{cases}
\label{eq:discomfort_reward}
\end{equation}



\subsection{Model}\label{subsec:model}

Our model builds upon the architecture described in \cite{EBCADRL}, extending it to incorporate information from the global path. The base architecture uses an attention mechanism to handle varying numbers of dynamic agents, while our extension integrates checkpoint features from the A* path to provide global guidance. We first describe the checkpoint feature extraction and extended self-state representation, then present the neural network architecture.

\paragraph{Checkpoint Feature Extraction.}
To incorporate global path information, we extract features from $K$ upcoming checkpoints along the A* path. At each time step, we must determine which checkpoints to use as input features. Let $m$ denote the index of the most recently visited checkpoint (with $m = -1$ initially). We consider two strategies for selecting the starting checkpoint index $s$.
\textit{Sequential selection}: The simplest approach sets $s = m + 1$, always selecting the $K$ checkpoints immediately following the last visited one. This ensures strict progression along the path and provides consistent, deterministic features.
\textit{Nearest-first selection}: This adaptive approach searches among all unvisited checkpoints with indices $i > m$ and selects the one nearest to the robot's current position as the starting point: $s = \arg\min_{i > m} \|c_i - \mathbf{p}\|_2$. This strategy provides better local guidance when the robot deviates from the planned path, as it focuses on the most immediately relevant checkpoint rather than one that may be far ahead on the path.

The sequential strategy offers simplicity and stability, while the nearest-first strategy improves adaptability at a modest computational cost. In both cases, the constraint $s > m$ prevents the robot from receiving features for already-visited checkpoints, ensuring forward progress along the global path.

Let $\mathcal{C} = \{c_s, c_{s+1}, \ldots, c_{s+K-1}\}$ denote the set of $K$ selected checkpoint positions. For each checkpoint $c_k = (c_x^k, c_y^k)$, we compute a feature vector in the robot-centric coordinate frame.

The robot-centric coordinate frame is defined such that the x-axis points from the robot's current position toward the goal. Let $\phi = \operatorname{atan2}(g_y - p_y, g_x - p_x)$ be the rotation angle, where $(p_x, p_y)$ is the robot position and $(g_x, g_y)$ is the goal position. For each checkpoint $c_k$, we compute:
\begin{equation}
\begin{aligned}
\delta_x^k &= c_x^k - p_x, \quad \delta_y^k = c_y^k - p_y, \\
\tilde{\delta}_x^k &= \delta_x^k \cos\phi + \delta_y^k \sin\phi, \\
\tilde{\delta}_y^k &= \delta_y^k \cos\phi - \delta_x^k \sin\phi, \\
d^k &= \sqrt{(\delta_x^k)^2 + (\delta_y^k)^2},
\end{aligned}
\label{eq:checkpoint_transform}
\end{equation}
where $(\tilde{\delta}_x^k, \tilde{\delta}_y^k)$ are the rotated offsets in the robot-centric frame and $d^k$ is the Euclidean distance to checkpoint $k$.

The checkpoint feature vector for each checkpoint is defined as:
\begin{equation}
\begin{aligned}
\mathbf{f}^k = [d^k, \tilde{\delta}_x^k, \tilde{\delta}_y^k, r_{\text{cp}}],
\end{aligned}
\label{eq:checkpoint_features}
\end{equation}
where $r_{\text{cp}}$ is the checkpoint radius. The complete checkpoint feature representation is the concatenation of all $K$ checkpoint features:
\begin{equation}
\begin{aligned}
\mathbf{f}_{\text{cp}} = [\mathbf{f}^1, \mathbf{f}^2, \ldots, \mathbf{f}^K].
\end{aligned}
\label{eq:checkpoint_concat}
\end{equation}

\paragraph{Extended Self-State.}
In the base EB-CADRL model, the robot's self-state in the rotated frame is represented as $\mathbf{s}^r = [d_g, v_{\text{pref}}, \theta, r, v_x, v_y]$, where $d_g$ is the distance to goal, $v_{\text{pref}}$ is the preferred speed, $\theta$ is the heading angle, $r$ is the robot radius, and $(v_x, v_y)$ is the velocity in the robot-centric frame.

In our HMP-DRL model, we extend the self-state by incorporating the checkpoint features:
\begin{equation}
\begin{aligned}
\tilde{\mathbf{s}}^r = [d_g, v_{\text{pref}}, \theta, r, v_x, v_y, \mathbf{f}_{\text{cp}}],
\end{aligned}
\label{eq:extended_self_state}
\end{equation}
resulting in an extended self-state of dimension $6 + 4K$. This extended self-state provides the model with explicit information about the global path, enabling it to make decisions that balance local collision avoidance with progress toward distant goals.

\paragraph{Entity State Embedding.}
We embed the extended robot state $\tilde{\mathbf{s}}^r$, along with the observable state of each entity $i$, into a fixed-length vector $g_i$ using a multi-layer perceptron (MLP):
\begin{equation}
\begin{aligned}
g_i = \phi_g(\tilde{\mathbf{s}}^r, \mathbf{s}^{io}; W_g),
\end{aligned}
\label{eq:mlp}
\end{equation}
where $\phi_g(\cdot)$ is an embedding function with rectified linear unit (ReLU) activations, and $W_g$ are the embedding weights. To account for different types of entities (e.g., adults, children, bicycles), we use one-hot encoding to represent each agent type. These encoded features are concatenated with the primary features of the agents and passed into the model.

The embedding vector $g_i$ is then fed into another MLP to obtain the pairwise interaction feature between the robot and entity $i$:
\begin{equation}
\begin{aligned}
h_i = \psi_h(g_i; W_h),
\end{aligned}
\label{eq:pairwise_interaction}
\end{equation}
where $\psi_h(\cdot)$ is a fully-connected layer with ReLU nonlinearity, and $W_h$ are the network weights.

\paragraph{Social Attentive Pooling.}
To handle varying numbers of entities, we use a social attentive pooling module \cite{SARL} which learns the relative importance of each neighbor and the collective impact of the crowd in a data-driven manner. Let $n$ denote the number of entities in the environment. The interaction embedding $g_i$ is transformed into an attention score $\alpha_i$ as follows:
\begin{equation} g_m = \frac{1}{n} \sum_{k=1}^n g_k, \qquad \alpha_i = \psi_\alpha(g_i, g_m; W_\alpha), \label{eq:attention_score2} \end{equation}
where $g_m$ is a fixed-length embedding vector obtained by mean pooling all individuals, $\psi_\alpha(\cdot)$ is an MLP with ReLU activations, and $W_\alpha$ are the weights.

Given the pairwise interaction vector $h_i$ and the corresponding attention score $\alpha_i$ for each neighbor $i$, the final representation of the crowd is a weighted linear combination of all pairs:
\begin{equation}
\begin{aligned}
c = \sum_{i=1}^n \operatorname{softmax}(\alpha_i) h_i,
\end{aligned}
\label{eq:crowd_representation}
\end{equation}
where $\operatorname{softmax}(\alpha_i) = \frac{e^{\alpha_i}}{\sum_{j=1}^n e^{\alpha_j}}.$

\paragraph{Value Network.}
Using the compact representation of the crowd $c$ and the extended self-state $\tilde{\mathbf{s}}^r$, we construct a planning module that estimates the state value $v$ for cooperative planning:
\begin{equation}
\begin{aligned}
v = f_v(\tilde{\mathbf{s}}^r, c; W_v),
\end{aligned}
\label{eq:state_value_v_for_cooperative_planning}
\end{equation}
where $f_v(\cdot)$ is an MLP with ReLU activations, and $W_v$ are the weights. We denote this value network as $V$.

In our model checkpoint features are incorporated into the extended self-state $\tilde{\mathbf{s}}^r$, which is then used throughout the entire network pipeline from the initial pairwise embedding $\phi_g$ through to the final value estimation $f_v$. This allows the network to learn representations that jointly consider both dynamic obstacle avoidance and global path following from the very first layer. During training, the network learns to balance following the global path (guided by checkpoint rewards) with avoiding dynamic obstacles (guided by collision and discomfort penalties).

\subsection{Algorithm HMP-DRL}\label{subsec:algorithm_v_learning}

The HMP-DRL algorithm integrates global path planning with local reinforcement learning through Parallel Deep V-learning. The training procedure runs multiple episodes in parallel, collects experience, and updates the value network $V$ using temporal-difference learning with experience replay and a fixed target network $\hat{V}$. The main training loop is described in Algorithm \ref{algorithm_1}, and the episode execution is described in Algorithm \ref{algorithm_2}.

\begin{minipage}{0.95\linewidth}

\begin{algorithm}[H]
\caption{HMP-DRL Training with Parallel Deep V-learning}\label{algorithm_1}
\begin{algorithmic}[1]
\State Run imitation learning with demonstration $D$, update value network $V$ with $D$
\State Initialize: target value network $\hat{V} \leftarrow V$, experience replay memory $E \leftarrow D$, parallel processes number $N$
\While{episode $<$ M}
	\State Initialize a multiprocessing Pool of size $N$
	\For{$k$ = episode, episode + $N$} in parallel:
	\State Execute episode $k$ using Algorithm \ref{algorithm_2}
	\EndFor
	\State Sample random minibatch tuples from $E$
	\State Update value network $V$ by gradient descent
	\If{episode $\mod$ UpdateInterval = 0}
		\State Update target value network $\hat{V} \leftarrow V$
	\EndIf
	\State Update $episode \leftarrow episode + N$
\EndWhile
\State \Return $V$
\end{algorithmic}
\end{algorithm}

\end{minipage}

\vspace{-1.5em}

\begin{minipage}{0.95\linewidth}
\begin{algorithm}[H]
\caption{HMP-DRL Episode Execution}\label{algorithm_2}
\begin{algorithmic}[1]
\State Initialize environment with random start $\mathbf{p}_0$ and goal $\mathbf{g}$
\State Compute global path $\mathcal{P} \leftarrow \text{A}^*(\text{OccupancyMap}, \mathbf{p}_0, \mathbf{g})$
\State Place checkpoints $\{c_1, \ldots, c_L\}$ along $\mathcal{P}$ at intervals $D$
\State Initialize: visited set $\mathcal{V} \leftarrow \emptyset$, buffer $S \leftarrow \emptyset$, $t \leftarrow 0$
\Repeat
	\State Set last visited checkpoint index: $m \leftarrow \max\{i : c_i \in \mathcal{V}\}$
	\State Select $K$ checkpoints: $\mathcal{C}_K \leftarrow \{c_{m+1}, \ldots, c_{\min(m+K, L)}\}$
	\State Compute checkpoint features $\mathbf{f}_{\text{cp}}$ using \eqref{eq:checkpoint_transform}--\eqref{eq:checkpoint_concat}
	\State Construct extended state $\tilde{\mathbf{s}}_t^r \leftarrow [d_g, v_{\text{pref}}, \theta, r, v_x, v_y, \mathbf{f}_{\text{cp}}]$
	\State Form joint state $\tilde{\mathbf{s}}_t^j \leftarrow [\tilde{\mathbf{s}}_t^r, \mathbf{s}_t^{1o}, \ldots, \mathbf{s}_t^{no}]$
	\State $\mathbf{a}_t \leftarrow \begin{cases} \text{RandomAction}(), & \epsilon \\ \arg \max_{\mathbf{a} \in \mathbf{A}} R(\tilde{\mathbf{s}}_t^j, \mathbf{a}) + \gamma^{\Delta t \cdot v_{\text{pref}}} V(\tilde{\mathbf{s}}_{t+\Delta t}^{j}), & 1-\epsilon \end{cases}$
	\State Execute $\mathbf{a}_t$, observe $r_t$ and next state
	\If{robot enters unvisited checkpoint $c_i$}
		\State $\mathcal{V} \leftarrow \mathcal{V} \cup \{c_i\}$, add $R_{\text{cp}}$ to $r_t$
	\EndIf
	\State Store $(\tilde{\mathbf{s}}_t^{j}, \mathbf{a}_t, r_t, \tilde{\mathbf{s}}_{t+\Delta t}^{j})$ in $S$
	\State $t \leftarrow t + \Delta t$
\Until{terminal state or $t \geq t_{\max}$}
\For{each $(\tilde{\mathbf{s}}_t^{j}, \mathbf{a}_t, r_t, \tilde{\mathbf{s}}_{t+\Delta t}^{j})$ in $S$}
	\State $\text{value}_t \leftarrow r_t + \gamma^{\Delta t \cdot v_{\text{pref}}} \hat{V}(\tilde{\mathbf{s}}_{t+\Delta t}^{j})$
	\State Store $(\tilde{\mathbf{s}}_t^{j}, \text{value}_t)$ in experience memory $E$
\EndFor
\end{algorithmic}
\end{algorithm}
\end{minipage}

At the beginning of each episode, we compute the global path $\mathcal{P}$ from the robot's start position to its goal using A* on the occupancy grid. We then place checkpoints $\{c_1, \ldots, c_L\}$ along $\mathcal{P}$ at intervals of distance $D$. At each time step, we select $K$ upcoming checkpoints, transform them into the robot-centric frame using \eqref{eq:checkpoint_transform}, and concatenate the resulting features $\mathbf{f}_{\text{cp}}$ with the base self-state to form $\tilde{\mathbf{s}}^r$ as defined in \eqref{eq:extended_self_state}. The extended joint state $\tilde{\mathbf{s}}_t^j = [\tilde{\mathbf{s}}_t^r, \mathbf{s}_t^{1o}, \ldots, \mathbf{s}_t^{no}]$ is then used for action selection and value estimation.

The global path is computed once per episode, avoiding the computational overhead of replanning at every step. Since each episode has a unique start-goal pair, the robot learns to follow diverse paths through the environment. The checkpoint features provide the local planner with global context, enabling navigation over long distances while the attention-based value network handles dynamic obstacle avoidance.

\subsection{Implementation Details}\label{subsec:impl_details}

We assume non-holonomic robot kinematics and use a unicycle model.
The action space consists of 81 discrete actions: a stop action (zero velocity) plus combinations of 5 speeds, exponentially distributed in the range (0, \(v_{\text{pref}}\)], and 16 directions, uniformly distributed in \([0, 2\pi)\).

For the global path representation, we use $K = 2$ checkpoints with radius $r_{\text{cp}} = 5$\,m and inter-checkpoint distance $D = 15$. Note that $K$ denotes the number of checkpoints fed to the model at each timestep, not the total number of checkpoints along the path, which depends on the path length and distance $D$. This results in $4K = 8$ checkpoint features appended to the base self-state, yielding an extended self-state dimension of $6 + 8 = 14$.

The neural network architecture consists of four multi-layer perceptrons (MLP). An MLP's structure is denoted by the tuple $(d_{\text{in}}, d_1, \dots, d_{\text{out}})$, which specifies the number of neurons in the input, hidden, and output layers, respectively. The dimensions of these MLP components are $\phi_g$: (25, 300, 200), $\psi_h$: (200, 200, 100), $\psi_{\alpha}$: (400, 200, 200, 1), and $f_v$: (114, 350, 250, 200, 1). 
The input dimension 25 to $\phi_g$ represents the joint state consisting of the extended robot state (14), agent's observable state (7), and entity type one-hot encoding (4). The 114 input dimensions to $f_v$ combine the extended self-state (14) with the attention-weighted feature representation of all agents (100).

We apply batch normalization \cite{BatchNorm} after each hidden linear layer and before the ReLU activation in all MLPs. This technique standardizes the inputs to each layer, which stabilizes training and allows for the use of higher learning rates.
Additionally, it accelerates convergence and provides a mild regularization effect that can help mitigate overfitting. In our experiments, batch normalization proved particularly beneficial for stability in the value-based reinforcement learning setting, where target values are non-stationary.

We implemented the policy in the deep learning framework PyTorch \cite{Pytorch} and trained it with a batch size of 100 (for each episode) using stochastic gradient descent. For imitation learning, we collected 3000 episodes of demonstration using ORCA and trained the policy 200 epochs with a learning rate of 0.01. For reinforcement learning, the learning rate is 0.001 and the discount factor $\gamma$ is 0.99. The exploration rate of the $\epsilon$-greedy policy decays linearly from 0.5 to 0.05 in the first 25000 episodes and stays at 0.05 for the remaining episodes.

In our experiments, all input features (robot state, entity states, and entity types) are obtained directly from the simulator. On a real robot, these would be provided by standard localization and perception systems \cite{SpringerHandbook}, whose implementation is beyond the scope of this work.

\section{Experiments}\label{sec:experiments}

\subsection{Urban Episode Generation}\label{episodes_generation}

We generate a large dataset of robot navigation episodes in a simulator, based on real urban geometry from a map, by rasterizing OpenStreetMap (OSM) data for a large European city \cite{OpenStreetMap}.
The data preparation pipeline generates map snapshots, semantically segments each snapshot into land use and infrastructure categories, collapses the semantics into a binary occupancy map for global planning, and finally selects start-goal pairs between which a path is guaranteed to exist. These maps are subsequently used as static layers, while dynamic agents are modeled during training and evaluation using ORCA \cite{ORCA} under the "invisible robot" assumption.

\paragraph{Map Snapshots from OSM.} We define a rectangular area centered on the center of the city and uniformly discretize it with a dense grid of viewpoints within a 29\,km $\times$ 38\,km region. For each viewpoint, we generate a high-zoom square raster snapshot (zoom~20) by stitching 256\,px map tiles from a self-hosted OSM tile server. Snapshots are rendered without overlay text labels to ensure a stable color palette and are saved as square images (2000\,px per side). This step follows the standard Web~Mercator projection tiling procedure: geographic coordinates are converted to tile coordinates at the target zoom level, neighboring tiles covering the desired area are retrieved, and the resulting mosaic is cropped to the required size.

Each snapshot is converted into a semantic raster by matching pixel colors to a fixed palette. We use the nearest color rule (Manhattan distance in RGB with a small threshold) to assign each image pixel to one of the following classes:
\begin{multicols}{3}
\begin{enumerate}
  \item Residential area
  \item Greenfield land
  \item Industrial area
  \item School/University/Hospital
  \item Allotments
  \item Parking
  \item Brownfield land
  \item Park / green zone
  \item Forest / park
  \item Road
  \item Commercial area
  \item Grass / meadow
  \item Sports pitch
  \item Water
  \item Building
\end{enumerate}
\end{multicols}

\paragraph{Binary Occupancy Map.} Next, the semantic raster is collapsed into a binary occupancy map encoding static traversability for the global planner. Specifically, classes corresponding to natural obstacles or impassable infrastructure (buildings, water, sports pitches, and grass/meadow) are marked as occupied, while all other classes (residential areas with sidewalk, soil with sidewalk, industrial areas, schools/universities/hospitals, allotments, parking, brownfield sites, parks/green zones, forests/parks, roads, and commercial areas) are marked as free.

\begin{figure}[ht]
\centering
\begin{subfigure}[t]{0.495\textwidth}
  \centering
  \includegraphics[width=\linewidth]{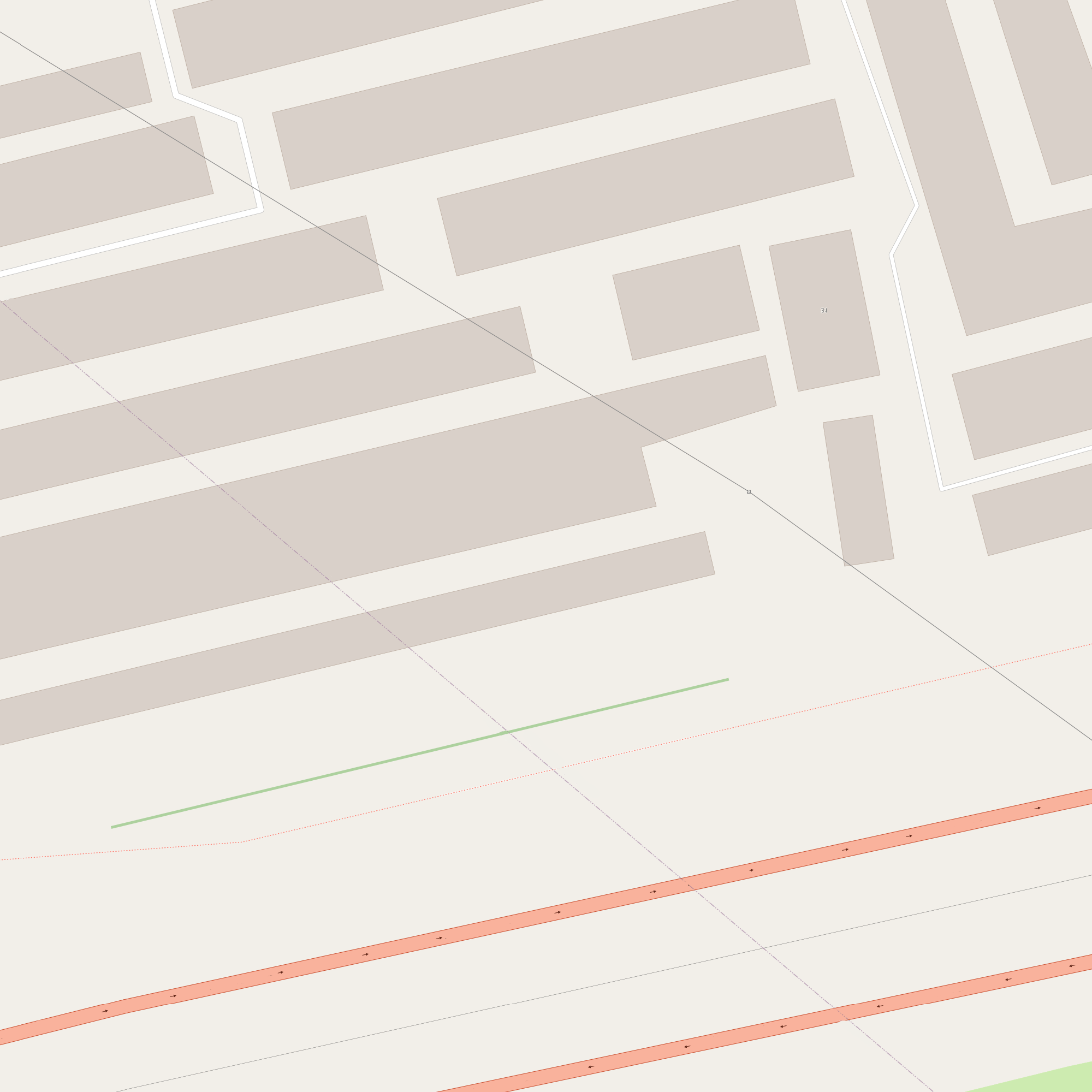}
  \caption{OSM map snapshot.}
  \label{fig:map_example_osm}
\end{subfigure}
\hfill
\begin{subfigure}[t]{0.495\textwidth}
  \centering
  \includegraphics[width=\linewidth]{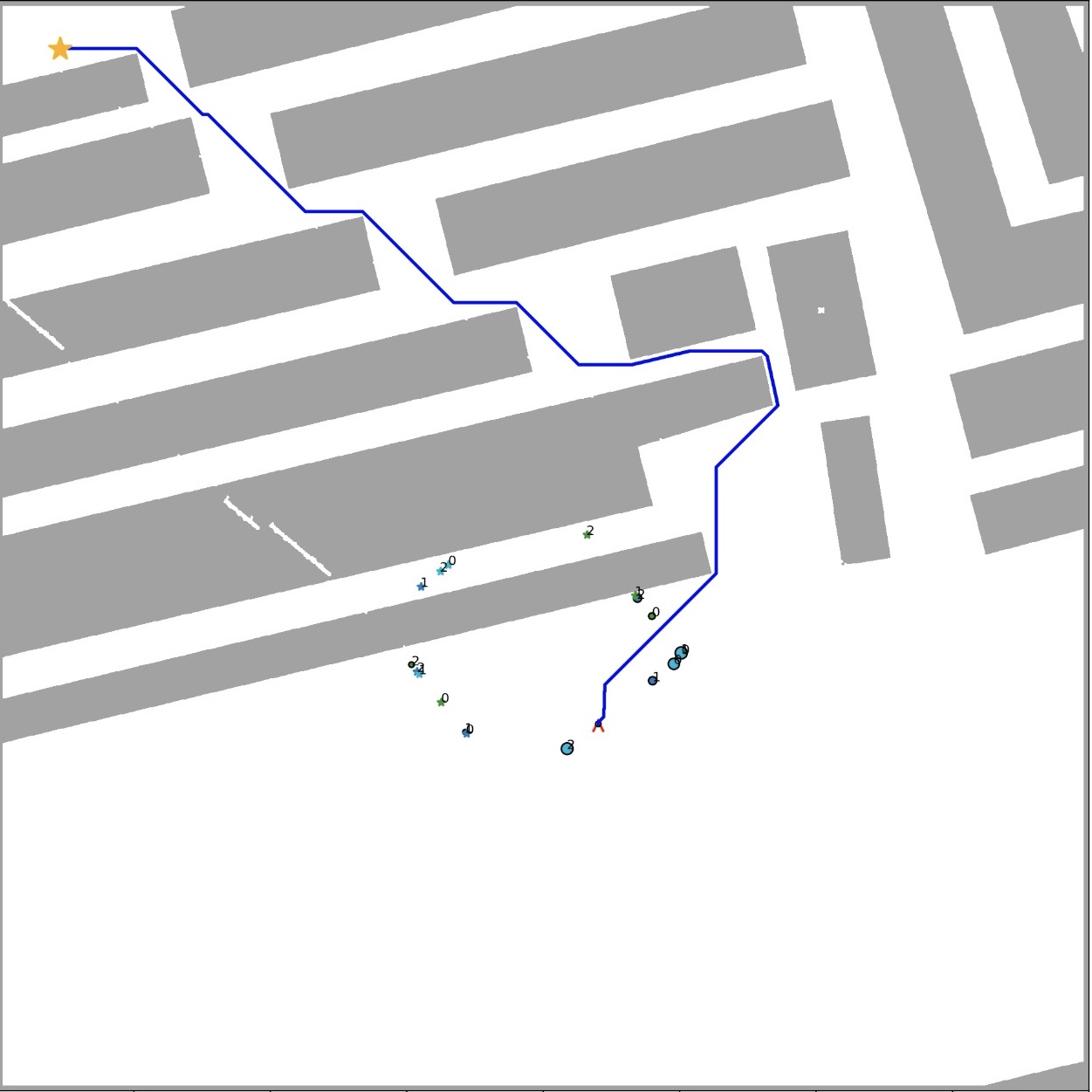}
  \caption{Binary occupancy map and global path.}
  \label{fig:map_example_grid}
\end{subfigure}
\caption{Example of map processing: (a) an OSM map snapshot from which occupancy maps were formed; (b) the corresponding binary occupancy map with a global path constructed using A*.}
\label{fig:map_examples}
\end{figure}

In Figs. \ref{fig:map_example_osm} and \ref{fig:map_example_grid}, we can see an example of the steps described above: an occupancy map is generated from an OSM city map, start and finish points are placed, and a global path is constructed.

\paragraph{Start-Goal Sampling.} On each binary map, we select between one and three start-goal pairs. The robot's footprint is modeled as a \(20\times20\)~pixel square; valid start and goal positions are those where the entire footprint lies completely within free space. Additionally, we require the goal to be sufficiently far from the start to yield non-trivial episodes: we select goals at a Euclidean distance of at least \(0.75\) of the map's larger dimension. To ensure dynamic feasibility at the global level, we inflate obstacles by half the footprint thickness and perform an 8-connected A* search (as described in Section~\ref{sec:new_approach}) on the inflated grid. Pairs are accepted only if a collision-free A* path exists; accepted paths are saved along with the start and goal.

\paragraph{Resulting Dataset.} Since we aim to train the robot to drive in sufficiently complex environments, we discard maps with no or very few static obstacles by introducing a minimum static occupancy fraction: maps with an obstacle percentage below the threshold (5\%) are skipped and unused.
Maps for which no valid start-goal pair satisfies the path existence constraint are also removed.

After filtering and removing unrealizable cases, we obtain 30,948 distinct maps. Sampling 1--3 start-goal pairs per map with a guaranteed path from start to finish yields 77,663 unique navigation episodes. These episodes serve as static occupancy maps within our simulator. We randomly split them into training, validation, and testing sets: for our main experiments, we use 60,000 episodes for training, 500 for validation, and 1,000 for testing. The remaining episodes are reserved for additional experiments as needed.

\subsection{Reward setup}\label{reward_setup}
In this subsection, we define the reward function for our experiments. The robot navigates in an environment with four entity types: adults, bicycles, children, and static obstacles. Each dynamic entity type has 4 instances, resulting in 12 dynamic agents per scenario. Static obstacles are defined by an occupancy map and vary by region.

\begin{equation}
\begin{aligned}
R_{\text{checkpoint}}(t) = 
\begin{cases}
0.3, & \text{if } \lVert \mathbf{p}_t - \mathbf{c}_k \rVert_2 < r_{\text{checkpoint}}  \text{ and $k$ is unvisited checkpoint} \\
0, & \text{otherwise}
\end{cases}
\end{aligned}
\label{eq:reward_checkpoint}
\end{equation}

\begin{equation}
R_{\text{success}} = 3
\label{eq:reward_success}
\end{equation}

We denote the entity types as follows: Adult ($A$), Bicycle ($B$), Child ($C$), and Obstacle ($O$). The penalties for collisions are defined as:

\begin{equation}
\begin{aligned}
R_{\text{coll}}(e) = 
\begin{cases}
-1.0, & \text{if } e = O \\
-1.5, & \text{if } e = A \\
-2.0, & \text{if } e = B \\
-2.5, & \text{if } e = C 
\end{cases}
\end{aligned}
\label{eq:penalties_for_collisions}
\end{equation}

We scale collision penalties according to unpredictability and harm potential. Static obstacles receive the lowest penalty (-1.0) since their positions are fixed. Adults incur a moderate penalty (-1.5) due to their predictable movement patterns. Cyclists warrant a higher penalty (-2.0) given their speed and fall risk. Collisions with children receive the maximum penalty (-2.5) as they are both unpredictable and highly vulnerable.

Let
$d_{\text{disc}}(e)$ be the discomfort distance (in meters) for entity $e$:
\begin{equation}
\begin{aligned}
d_{\text{disc}}(e) = 
\begin{cases}
0.1, & \text{if } e = A \\
0.2, & \text{if } e = B \\
0.2, & \text{if } e = C
\end{cases}
\end{aligned}
\label{eq:discomfort_distance}
\end{equation}

Let
$p_{\text{disc}}(e)$ be the discomfort penalty factor for entity $e$:
\begin{equation}
\begin{aligned}
p_{\text{disc}}(e) = 
\begin{cases}
0.0, & \text{if } e = O \\
0.5, & \text{if } e = A \\
1.0, & \text{if } e = B \\
1.0, & \text{if } e = C
\end{cases}
\end{aligned}
\label{eq:discomfort_penalty}
\end{equation}

We omit $d_{\text{disc}}(O)$ because static obstacles cannot experience discomfort. However, if the robot should maintain distance from static obstacles, one can define $d_{\text{disc}}(O)$ and set $p_{\text{disc}}(O) > 0$.

\subsection{Simulation Setup}\label{sec42}
We conduct experiments using our simulator designed for outdoor mobile robots operating in urban environments, such as sidewalk delivery robots, autonomous cleaners, and security patrol robots. The simulator models realistic pedestrian traffic and urban geometry based on the maps described in Section~\ref{episodes_generation}.

We compare three navigation methods: SARL \cite{SARL}, EB-CADRL \cite{EBCADRL}, and our HMP-DRL.

Simulated agents (adults, children, bicyclists) are controlled by Optimal Reciprocal Collision Avoidance (ORCA) \cite{ORCA}. To introduce diversity, each agent's size and speed are drawn from a uniform distribution specific to its type. The distributions are designed to overlap, for instance, some children may be faster than some adults, making it impossible to determine an agent's type from its size and speed alone.

\paragraph{Dynamic agent spawning.}
In our simulator we use new dynamic agents during an episode (``dynamic spawn'') to maintain traffic around the robot.
In our setup, dynamic spawning is enabled and executed every 80 simulation steps (with $\Delta t = 0.25$\,s, this corresponds to 20\,s).
At each spawn event, we add 4 adults, 4 bicyclists, and 4 children, and assign them a lifetime of 80 steps. Agents whose age exceeds the lifetime are removed from the scene.
New agents are sampled from a square region of width 40\,m, positioned and oriented relative to the robot: we compute the unit direction vector from the robot to its goal and place the square center ahead of the robot by half of the square width along this direction. The square yaw is aligned with the same robot-to-goal direction.
Start and goal positions for spawned agents are sampled on opposite edges of this oriented square (square-crossing), with rejection sampling to avoid immediate collisions with the robot, other agents, and static obstacles.

All evaluations use an ``invisible robot'' setting, where other agents do not perceive or react to the robot.
This tests the robot's collision avoidance capabilities in a worst-case scenario that is also realistic, as small robots are often overlooked by people.

\begin{figure*}[!htbp]
	\center{\includegraphics[width=0.99\linewidth]{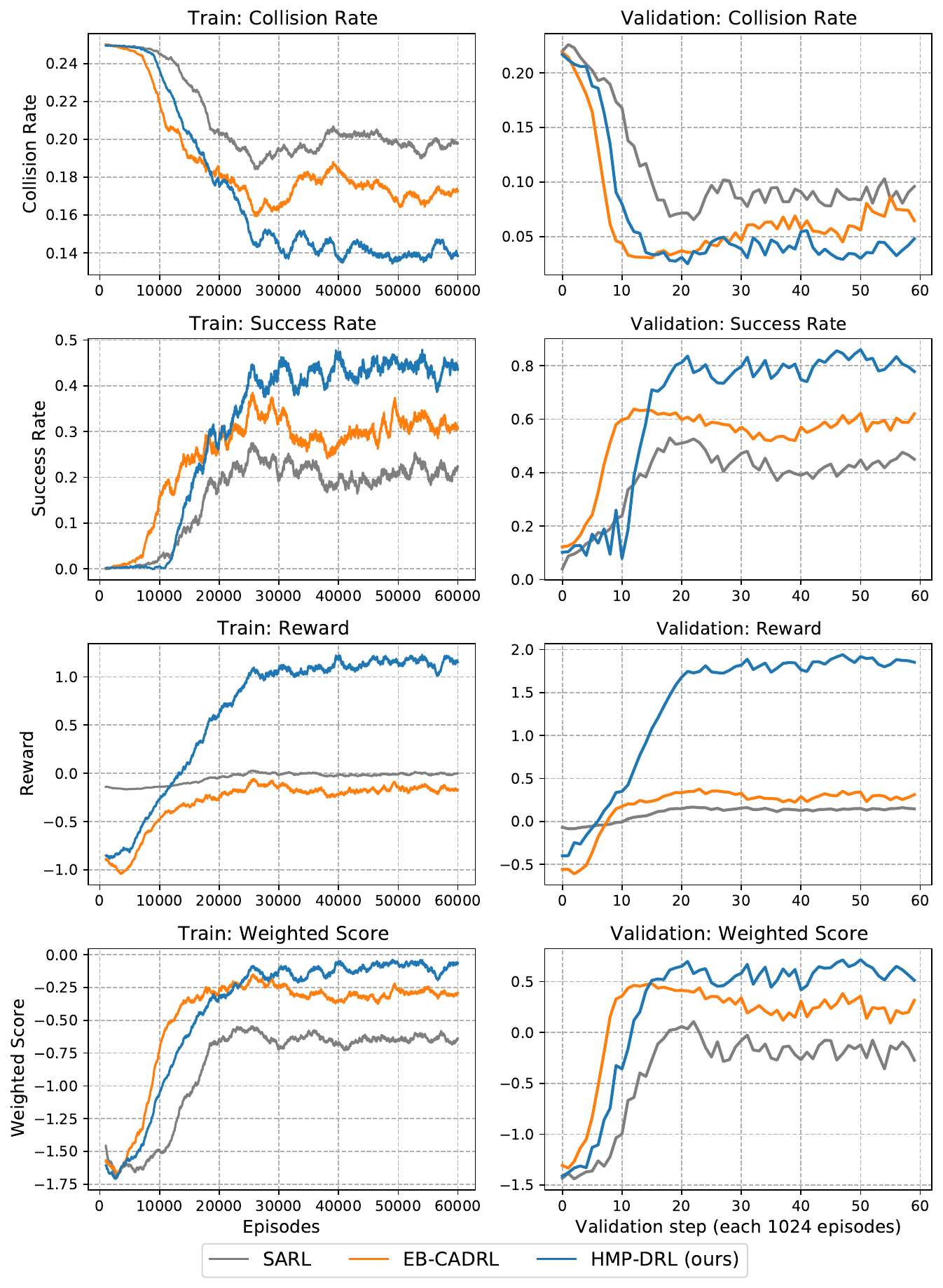}} 
	\caption{Collision rate, success rate, reward, weighted score metrics on the training and validation data. The validation step is performed every 1024 episodes.}
	\label{figure_sr_reward}
\end{figure*}

Models are trained for 3000 episodes with imitation learning, followed by 60000 episodes using the algorithm "HMP-DRL Training with Parallel Deep V-learning" \ref{algorithm_1}. During training, we validate the models every 1024 episodes on a 500-episode validation set. For the final comparison, we select the best-performing model for each method based on its validation performance using the weighted score metric and evaluate it on a 1,000-episode test set.

\subsection{Results}\label{subsec:results}

Figure~\ref{figure_sr_reward} shows training and validation curves.

All methods converge within 60{,}000 RL training episodes (after the imitation learning warm start).

Across both training and validation, HMP-DRL achieves consistently higher success rate and weighted score while maintaining lower collision rates than SARL and EB-CADRL.

To compare safety in a way that reflects different severity of collisions, we use the Weighted Score $(WS)$ metric:
\begin{equation}
\begin{aligned}
WS &= SR - CR(A) - 4.0 \cdot CR(C) - 2.0 \cdot CR(B) - 0.5 \cdot CR(O) .
\end{aligned}
\label{eq:weighted_score}
\end{equation}

Here, \(SR\) is the success rate and \(CR(\cdot)\) is the collision rate with adults (A), children (C), bicycles (B), and static obstacles (O). Coefficients are chosen to penalize collisions with 
more vulnerable entities more heavily and can be tuned to match 
application-specific safety priorities.

Since each method uses a different reward definition, episode returns are not directly comparable across methods and are used only to monitor within-method learning progress.

\begin{table*}[!t]
\centering
\setlength{\tabcolsep}{4pt}
\caption{Metrics on test data with 12 agents.}
\label{test_table}
\begin{tabular}{lccc}
\hline
	Method & SARL  & EB-CADRL & HMP-DRL \\
	SR     & 0.524 & 0.649    & \bfseries 0.836     \\
	CR     & 0.289 & 0.128    & \bfseries 0.123     \\
	CR(A)  & 0.05  & \bfseries 0.019    & \bfseries 0.019     \\
	CR(B)  & 0.155 & 0.047    & \bfseries 0.044     \\
	CR(C)  & 0.022 & 0.015    & \bfseries 0.004     \\
	CR(O)  & 0.062 & \bfseries 0.047    & 0.056     \\
	Time   & 78.36 & 93.42    & \bfseries 72.03     \\
	DD(A)  & 0.154 & 0.147    & \bfseries 0.166     \\
	DD(B)  & 0.144 & 0.181    & \bfseries 0.183     \\
	DD(C)  & 0.153 & 0.184    & \bfseries 0.209     \\
	WS     & 0.045 & 0.4525   & \bfseries 0.685     \\
\hline
\end{tabular}
\end{table*}

Table~\ref{test_table} summarizes evaluation on the 1,000-episode test set. We report: \textit{SR} (success rate); \textit{CR} (overall collision rate); \textit{CR(A), CR(B), CR(C), CR(O)} (collision rates with adults, bicycles, children, and static obstacles); \textit{Time} (average time to reach the goal); and \textit{DD(A), DD(B), DD(C)} (mean robot--agent distance during ``danger'' interactions for adults, bicycles, and children, where danger is defined as distance \(< 30\,\text{cm}\)).

HMP-DRL achieves a substantially higher success rate (0.836) than EB-CADRL (0.649) and SARL (0.524), while also reducing the overall collision rate (0.123 vs.\ 0.128 and 0.289). The largest safety gain is in collisions with children: \(CR(C)\) drops to 0.004 for HMP-DRL (vs.\ 0.015 for EB-CADRL and 0.022 for SARL), which strongly improves the weighted score (0.685 vs.\ 0.4525 and 0.045). HMP-DRL also reaches the goal faster on average (72.03\,s vs.\ 93.42\,s and 78.36\,s).

\begin{table*}[!t]
\centering
\setlength{\tabcolsep}{4pt}
\caption{Metrics on test data with 3 agents (1 adult, 1 bicycle, 1 child).}
\label{test_table_agents_3}
\begin{tabular}{lccc}
\hline
	Method & SARL   & EB-CADRL & HMP-DRL \\
	SR     & 0.719  & 0.712    & \bfseries 0.936     \\
	CR     & 0.054  & 0.034    & \bfseries 0.017     \\
	CR(A)  & 0.009  & 0.003    & \bfseries 0.001     \\
	CR(B)  & 0.018  & 0.015    & \bfseries 0.007     \\
	CR(C)  & 0.004  & \bfseries 0.0      & \bfseries 0.0       \\
	CR(O)  & 0.023  & 0.016    & \bfseries 0.009     \\
	Time   & 89.72  & 98.89    & \bfseries 72.84     \\
	DD(A)  & 0.15   & 0.166    & \bfseries 0.182     \\
	DD(B)  & 0.151  & \bfseries 0.169    & 0.15      \\
	DD(C)  & 0.157  & \bfseries 0.249    & 0.208     \\
	WS     & 0.6465 & 0.671    & \bfseries 0.9165    \\
\hline
\end{tabular}
\end{table*}

Table~\ref{test_table_agents_3} reports results in a simpler scenario with 3 dynamic agents (1 adult, 1 bicycle, 1 child), which reduces the influence of dense surrounding traffic. In this setting, HMP-DRL further improves performance, achieving \(SR=0.936\) and \(CR=0.017\), with near-zero collisions with vulnerable agents (\(CR(A)=0.001\), \(CR(C)=0.0\)). Consequently, it attains the highest weighted score (0.9165) and the lowest average time to reach the goal (72.84\,s).

Overall, HMP-DRL performs best on the main metrics: it achieves the highest success rate (SR) and weighted score (WS), and the lowest collision rate (CR), while maintaining comparable safety margins during close interactions.

\subsection{Ablation Experiments}\label{sec43}

\begin{table}[ht!]
\centering
\caption{Ablation study: effect of checkpoint reward shaping (\(R_{\text{cp}}\)) on HMP-DRL.}
\label{ablation_table}
\begin{tabular}{lcc}
\hline
	Method & With checkpoint reward & Without checkpoint reward \\
	       & (\(R_{\text{cp}}=0.3\)) & (\(R_{\text{cp}}=0.0\)) \\
	SR     & \bfseries 0.836                     & 0.615                   \\
	CR     & \bfseries 0.123                     & 0.13                    \\
	CR(A)  & 0.019                     & \bfseries 0.018                   \\
	CR(B)  & \bfseries 0.044                     & 0.054                   \\
	CR(C)  & \bfseries 0.004                     & 0.008                   \\
	CR(O)  & 0.056                     & \bfseries 0.05                    \\
	Time   & \bfseries 72.03                     & 79.59                   \\
	DD(A)  & \bfseries 0.166                     & 0.153                   \\
	DD(B)  & \bfseries 0.183                     & 0.182                   \\
	DD(C)  & \bfseries 0.209                     & 0.2                     \\
	WS     & \bfseries 0.685                     & 0.432                   \\
\hline
\end{tabular}
\end{table}

We ablate the checkpoint-based reward shaping term (Section~\ref{subsec:reward_function}, Eq.~\eqref{eq:checkpoint_reward}), which provides intermediate progress feedback along the global A* path via checkpoints. We train two HMP-DRL policies from scratch with identical architecture, datasets, and hyperparameters; the only difference is the checkpoint reward magnitude \(R_{\text{cp}}\). In the first variant we use the default setting \(R_{\text{cp}}=0.3\), while in the second we disable checkpoint shaping by setting \(R_{\text{cp}}=0.0\). Both policies are evaluated on the same 1,000-episode test set as in Table~\ref{test_table}, and we report the same metrics.

The results are summarized in Table~\ref{ablation_table}. Enabling checkpoint rewards substantially improves navigation performance: the success rate increases from 0.615 to 0.836 and the weighted score increases from 0.432 to 0.685. The policy also reaches the goal faster on average (72.03\,s vs.\ 79.59\,s), indicating more efficient progress and fewer stalled trajectories. Safety improves primarily for the most challenging dynamic entities: collisions with bicycles decrease (0.054 \(\rightarrow\) 0.044) and collisions with children are halved (0.008 \(\rightarrow\) 0.004). We also observe larger mean danger distances \(DD(\cdot)\), suggesting the checkpoint-shaped policy maintains slightly larger safety margins during close interactions. While \(CR(A)\) and \(CR(O)\) change only marginally (and in opposite directions), the overall effect of checkpoint shaping is strongly positive across the key metrics.

These findings support our design choice: sparse checkpoint feedback aligns the local policy with the global plan, improving all main metrics.

\section{Conclusion}\label{sec:conclusions}

In this work, we proposed \emph{Hybrid Motion Planning with Deep Reinforcement Learning} (HMP-DRL), a hybrid navigation approach that couples a graph-based global planner with a local DRL policy. The global level computes global path and converts it into a sequence of distance-spaced checkpoints. The local level extends an attention-based, entity-aware value network with checkpoint features and trains it with reward shaping that encourages progress along the route while enforcing type-dependent safety margins and collision penalties for different entities.

Experiments in a large-scale urban simulator derived from OpenStreetMap, with dense ORCA-controlled traffic, show that HMP-DRL improves both reliability and safety. On the main test set, it achieves a success rate of 0.836 (vs.\ 0.649 for EB-CADRL and 0.524 for SARL) with a lower overall collision rate of 0.123 and faster traversal time (72.03\,s on average). The largest gain is for the most vulnerable entity type: collisions with children drop to 0.004, yielding the best weighted safety score (0.685). An ablation study further confirms the role of route guidance: disabling checkpoint rewards reduces the success rate from 0.836 to 0.615 and the weighted score from 0.685 to 0.432.

Future research directions include integrating 
dynamic occupancy maps with online graph 
re-planning, adding additional entity types (e.g., 
cars and motorcycles), training policies for complex 
scenarios—such as road crossings with dynamic 
vehicles, and real-world testing.

\bibliographystyle{unsrtnat}
\bibliography{bibliography.bib}

\end{document}